\documentclass[sigconf]{acmart}

\AtBeginDocument{%
  \providecommand\BibTeX{{%
    \normalfont B\kern-0.5em{\scshape i\kern-0.25em b}\kern-0.8em\TeX}}}

\setcopyright{acmcopyright}
\copyrightyear{2022}
\acmYear{2022}
\acmDOI{10.1145/3503161.3547868}

\acmConference[MM'2022]{ACM MULTIMEDIA CONFERENCE 2022}{October 10--14, 2022}{Lisbon, Portugal}
\acmPrice{15.00}
\acmISBN{978-1-4503-9203-7/22/10}

\acmSubmissionID{495}

\usepackage{multirow}
\usepackage{enumitem}
\usepackage{bbding}
\usepackage{url}
\begin{document}

\title{Modality-Aware Contrastive Instance Learning with Self-Distillation for Weakly-Supervised Audio-Visual Violence Detection}

\author{Jiashuo Yu\textsuperscript{1,*}\authornotemark[1], Jinyu Liu\textsuperscript{1,*}\authornotemark[1], Ying Cheng\textsuperscript{2}, Rui Feng\textsuperscript{1,2,3,$\dagger$}\authornotemark[2], Yuejie Zhang\textsuperscript{1,3,$\dagger$}\authornotemark[2]}

\affiliation{
\textsuperscript{1}School of Computer Science, Shanghai Key Laboratory of Intelligent Information Processing, Fudan University, China
\\\textsuperscript{2}Academy for Engineering and Technology, Fudan University, China
\\\textsuperscript{3}Shanghai Collaborative Innovation Center of Intelligent Visual Computing, Fudan University, China
\country{}}

\email{{jsyu19,jinyuliu20,chengy18,fengrui,yjzhang}@fudan.edu.cn}

\begin{abstract}
Weakly-supervised audio-visual violence detection aims to distinguish snippets containing multimodal violence events with video-level labels. Many prior works perform audio-visual integration and interaction in an early or intermediate manner, yet overlooking the modality heterogeneousness over the weakly-supervised setting. In this paper, we analyze the \textbf{modality asynchrony} and \textbf{undifferentiated instances} phenomena of the multiple instance learning (MIL) procedure, and further investigate its negative impact on weakly-supervised audio-visual learning. To address these issues, we propose a modality-aware contrastive instance learning with self-distillation (MACIL-SD) strategy . Specifically, we leverage a lightweight two-stream network to generate audio and visual bags, in which unimodal background, violent, and normal instances are clustered into semi-bags in an unsupervised way. Then audio and visual violent semi-bag representations are assembled as positive pairs, and violent semi-bags are combined with background and normal instances in the opposite modality as contrastive negative pairs. Furthermore, a self-distillation module is applied to transfer unimodal visual knowledge to the audio-visual model, which alleviates noises and closes the semantic gap between unimodal and multimodal features. Experiments show that our framework outperforms previous methods with lower complexity on the large-scale XD-Violence dataset. Results also demonstrate that our proposed approach can be used as plug-in modules to enhance other networks. Codes are available at \color{magenta}{\url{https://github.com/JustinYuu/MACIL_SD}.}
\end{abstract}


\maketitle

\renewcommand{\thefootnote}{\fnsymbol{footnote}}
\footnotetext[1]{Equal contribution.}
\footnotetext[2]{Corresponding authors.}

\begin{figure}[htp]
\centering
\includegraphics[width=0.47\textwidth]{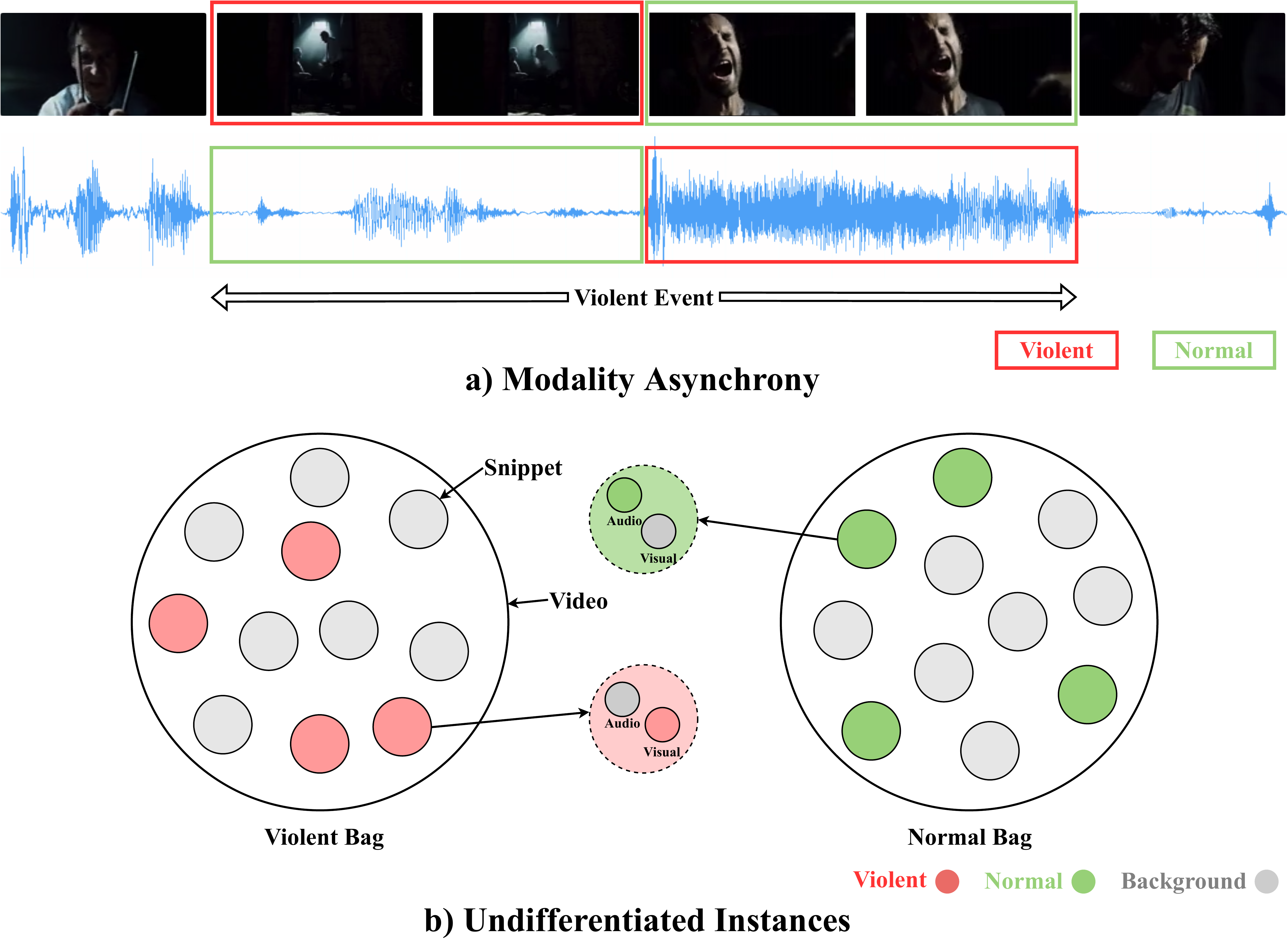}
\vspace{-2mm}
\caption{a) An example of the modality asynchrony. During the violent event \textit{abuse}, the abuser first hits the victim, where the violent message is reflected in the visual modality. Then the scream of the victim occurs, indicating the auditory violence information. b) The illustration of the undifferentiated instances. In each bag, violent cues are distributed in some instances while others contain background noises, and the discrepancy between normal segments and background noises also exists. We argue that adding additional constraints could enhance model discrimination.}
\label{figure1}
\end{figure}

\section{Introduction}

Recent years have witnessed the extension of violence detection from a pure vision task~\cite{bermejo2011violence, deniz2014fast, zhang2019temporal, peixoto2019toward, zhang2016new, khan2019cover, feng2021mist, tian2021weakly, wu2021learning, ristea2021self, li2022self} to an audio-visual multimodal problem~\cite{wu2020not, peixoto2020multimodal, pang2021violence}, for which the corresponding auditory content supplements fine-grained violent cues. Despite numerous modality fusion and interaction methods have shown promising results, the modality discrepancy of the multiple instance learning (MIL)~\cite{maron1997framework} framework under the weakly-supervised setting remains to be explored.

To alleviate the appetite for fine-labeled data, MIL is widely adopted for the weakly-supervised violence detection, where the output of each video sequence is formed into a bag containing multiple snippet-level instances. In the audio-visual scenarios, all prior works share a general scheme that regards each audio-visual snippet as an integral instance and averaging the top-\textit{K} audio-visual logits as the final video-level scores. However, we analyze that this formula suffers from two defects: \textbf{modality asynchrony} and \textbf{undifferentiated instances}. Modality asynchrony indicates the temporal inconsistency between auditory and visual violence cues. Taking the typical violent event \textit{abuse} in Figure~\ref{figure1}(a) as an example, when the abuser hits the victim, the scream occurs afterward, and the entire procedure is regarded as a violent event. In this situation, scenes in part of the visual modality (2nd-3rd snippets) and audio modality (4th-5th snippets) contain violent clues. We argue that directly leveraging an audio-visual pair as an instance could introduce data noise to the video-level optimization. The other defect we discovered is undifferentiated instances, that is, picking the top-
\textit{K} instances for optimization results in numerous disengaged instances. As shown in Figure~\ref{figure1}(b), in a sequence of violent videos, the violent event can be reflected in some audio/visual instances. In contrast, others contain irrelevant elements such as background noises. On the contrary, in the videos of normal events, a few snippets contain elements of normal events, while others include background information. In this case, the \textit{K}-max activation abandons the instances containing background elements, and the discrepancy between violent and normal instances is not explicitly revealed. To this end, we argue that adding contrastive constraints among the violent, normal, and background instances could contribute to the discrimination toward violent content.

Driven by preliminary analysis, we propose a simple yet effective framework constructed by modality-aware contrastive instance learning (MA-CIL) and self-distillation (SD) module. To address the modality asynchrony, we form the unimodal bags apart from the original audio-visual bags, compute unimodal logits, and cluster embeddings of top-\textit{K} and bottom-\textit{K} unimodal instances as semi-bags. To differentiate instances, we propose a modality-aware contrastive-based method. In detail, the audio and visual violent semi-bags are constructed as the positive pairs, while the violent semi-bags are assembled with embeddings of instances in the background and normal semi-bags as negative pairs. Furthermore, a self-distillation module is applied to distill unimodal knowledge to the audio-visual model, which closes the semantic gap between modalities and alleviates the data noise introduced by the abundant cross-modality interactions. In summary, our contributions are as follows:
\begin{itemize}[leftmargin=*]
\item We analyze the modality asynchrony and undifferentiated instances phenomena of the widely-used MIL framework in audio-visual scenarios, further elaborating their disadvantages for the weakly-supervised audio-visual violence detection.
\item We propose a modality-aware contrastive instance learning with self-distillation framework to introduce feature discrimination and alleviate modality noise.
\item Equipped with a lightweight network, our framework outperforms the state-of-the-art methods on the XD-Violence dataset, and our model also shows the generalizability as plug-in modules. 
\end{itemize}

\section{Related Works}

\subsection{Weakly-Supervised Violence Detection} 

Weakly-supervised violence detection requires identifying violent snippets under video-level labels, where the MIL~\cite{maron1997framework} framework is widely used for denoising irrelevant information. Some previous works~\cite{bermejo2011violence, deniz2014fast, zhang2019temporal, peixoto2019toward, zhang2016new, khan2019cover, wu2021learning, ristea2021self, feng2021mist} regard violence detection as a pure vision task and leverage CNN-based networks to encode visual features. Among these methods, various feature integration and amelioration methods are proposed to enhance the robustness of MIL. Tian et al.~\cite{tian2021weakly} propose RTFM, a robust temporal feature magnitude learning method to refine the capacity of recognizing positive instances. Li et al.~\cite{li2022self} design a Transformer~\cite{vaswani2017attention}-based multi-sequence learning network to reduce the probability of instance selection errors. However, these models neglect the corresponding auditory information as well as the cross-modality interactions, thereby restricting the performance of violence prediction.

Recently, Wu et al.~\cite{wu2020not} curate a large-scale audio-visual dataset XD-Violence and establish an audio-visual benchmark. However, they integrate audio and visual features in an early fusion way, thereby limiting further inter-modality interactions. To facilitate multimodal fusion, Pang et al.~\cite{pang2021violence} propose an attention-based network to adaptively integrate audio and visual features with mutual learning module in an intermediate manner. Different from prior methods, we perform inter-modality interactions via a lightweight two-stream network and conduct discriminative multimodal learning via modality-aware contrast and self-distillation.

\subsection{Contrastive Learning} 
Contrastive learning is formulated by contrasting positive pairs against negative pairs without data supervisory. In the unimodal field, several visual methods~\cite{chen2020simple, jinyuliu2022, he2020momentum, grill2020bootstrap} leverage the augmentation of visual data as a contrast to increase model discrimination. Furthermore, some natural language processing methods utilize the token- and sentence-level contrasts to enhance the performance of pre-trained models~\cite{clark2020electra, su2021tacl} and supervised tasks~\cite{peng2020learning, das2021container}. For the multimodal fields, some works introduce modality-aware contrasts to vision-language tasks, such as image captioning~\cite{dai2017contrastive, wang2022distinctive}, visual question answering~\cite{chen2021counterfactual, whitehead2021separating}, and representation learning~\cite{shi2020contrastive, li2020unimo, yang2022vision, wen2021cookie}. Moreover, recent literature~\cite{arandjelovic2017look, owens2018audio, korbar2018cooperative, arandjelovic2018objects, cheng2020look, ma2021active, morgado2021audio, morgado2021robust} utilizes the temporal consistency of audio-visual streams as contrastive pretext tasks to learn robust audio-visual representations. Based on existing instance-level contrastive frameworks~\cite{chen2021cil, wu2018unsupervised}, we put forward the concept of semi-bags and leverage the cross-modality contrast to obtain model discrimination.

\begin{figure*}[t]
\centering
\includegraphics[width=0.96\textwidth]{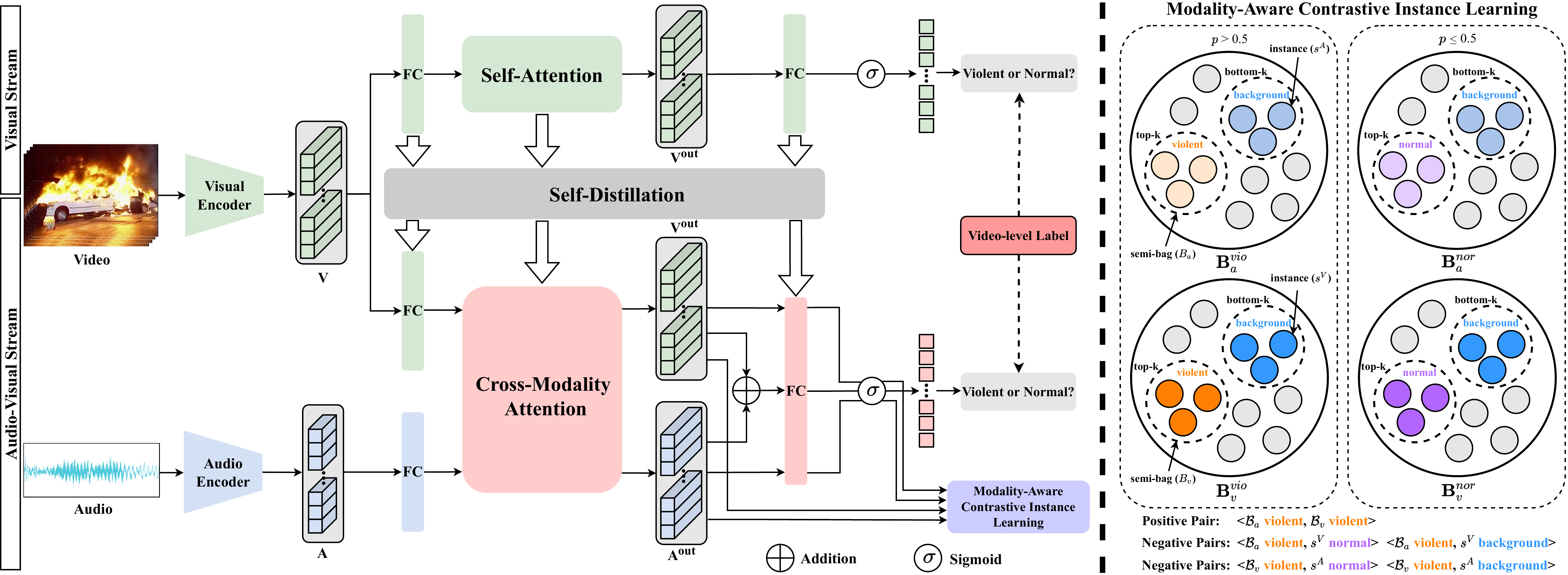}
\vspace{-2mm}
\caption{An illustration of our proposed Modality-Aware Contrastive Instance Learning with Self-Distillation framework. Our approach consists of three parts: the lightweight two-stream network, modality-aware contrastive learning (MA-CIL), and self-distillation (SD) module. Taking audio and visual features extracted from pretrained networks as inputs, we design a simple yet effective attention-based network to perform audio-visual interaction. Then a modality-aware contrasting-based method is used to cluster instances of different types into several semi-bags and further obtain model discrimination. Finally, a self-distillation module is deployed to transfer visual knowledge to our audio-visual network, aiming to alleviate modality noise and close the semantic gap between unimodal and multimodal features. The entire framework is trained jointly in a weakly supervised manner, and we adopt the multiple instance learning (MIL) strategy for optimization.}
\vspace{-2mm}
\label{figure2}
\end{figure*}

\subsection{Cross-Modality Knowledge Distillation} 
Knowledge distillation is first proposed to transfer knowledge from large-scale architectures to lightweight models~\cite{hinton2015distilling, bucilu2006model}. However, the cross-modality distillation aims to transfer unimodal knowledge to multimodal models for alleviating the semantic gap between modalities. Several methods~\cite{garcia2018modality, hoffman2016learning} distill depth features to the RGB representations via hallucination networks to address the modality missing and noisy phenomena. Chen et al.~\cite{chen2021distilling} propose an audio-visual distillation strategy, which learns the compositional embedding and transfers knowledge across semantic-uncorrelated modalities. Recently, Multimodal Knowledge Expansion~\cite{xue2021multimodal} is proposed as a two-stage distillation strategy, which transfers knowledge from unimodal teacher networks to the multimodal student network by generating pseudo labels. Inspired by the methodology of self-distillation~\cite{fang2021seed, caron2021emerging, chen2020big, xie2020self, tian2019contrastive1, chen2021wasserstein}, we propose the parameter integration paradigm to transfer visual knowledge to our audio-visual model via two similar lightweight networks, which reduces the modality noise and benefits robust audio-visual representation.

\section{Preliminaries}

Given an audio-visual video sequence $S = (S^A, S^V)$, where $S^A$ is the audio channel, and $S^V$ denotes the visual channel, the entire sequence is divided into $T$ non-overlapping segments $\{s_t^A, s_t^V\}_{t=1}^N$. For an audio-visual pair ($s_t^A, s_t^V$), weakly-supervised violence detection task requires to distinguish whether it contains violent events via an event relevance label $y_t \in \{0, 1\}$, where $y_t = 1$ means at least one modality in the current segment includes violent cues. In the training phase, only video-level labels $y$ are available for optimization. Hence, a general scheme is to utilize the multiple instance learning (MIL) procedure to satisfy the weak supervision.

In the MIL framework, each video sequence $S$ is regarded as a bag, and video segments $\{s_t^A, s_t^V\}_{t=1}^N$ are taken as instances. Then instances are aggregated via a specific feature-level/score-level pooling method to generate video-level predictions $p$. In this paper, we utilize the \textit{K}-max activation with average pooling rather than attention-based methods~\cite{nguyen2018weakly, tian2020unified} and global pooling~\cite{zhang2019temporal, sultani2018real} as the aggregation function. To be specific, given the audio and visual feature $f_a, f_v$ extracted by CNN networks, we use a multimodal network to generate unimodal logits $l_{a}, l_{v}$, and audio-visual logits $l_{av}$. The embeddings of audio and visual instances are symbolized as $h_a$ and $h_v$. Then we average $K$ maximum logits and use the sigmoid activation to generate the video-level prediction $p$. Due to the additional constraint of our proposed contrastive learning method, we define the unimodal bags $\mathbf{B}_a, \mathbf{B}_v$. In each unimodal bag, instances are clustered into several semi-bags $B_m, m\in \{a, v\}$ based on their intrinsic characteristics, and the corresponding semi-bag representations are noted as $\mathcal{B}_m, m\in \{a, v\}$.

\section{Methodology}

Our proposed framework consists of three parts, a lightweight two-stream network, modality-aware contrastive instance learning (MA-CIL), and the self-distillation (SD) module. An illustration of our framework shown in Figure~\ref{figure2} is detailed as follows. 

\subsection{Two-Stream Network}

Considering prior methods suffer from the parameter redundancy of the large-scale networks, we design an encoder-agnostic lightweight architecture to achieve feature aggregation and modality interaction. Taking the visual and auditory feature $f_v$, $f_a$ extracted by pre-trained networks (e.g., I3D and VGGish for visual and audio features, respectively) as input, our proposed network consists of three parts, linear layers to keep the dimension of input features identical, cross-modality attention layer to perform inter-modality interactions, and MIL module for the weakly-supervised training. Among these modules, the cross-modality attention layer is ameliorated from the encoder part of Transformer~\cite{vaswani2017attention}, which includes the multi-head self-attention, feed-forward layer, residual connection~\cite{he2016deep}, and layer normalization~\cite{ba2016layer}. In the raw self-attention block, features are projected by three different parameter matrices as query, key, and value vectors, respectively. Then the scale dot-product attention score is computed by $att(q, k, v) = softmax(\frac{qk^T}{\sqrt{d_m}})v$, where $q, k, v$ denotes the query, key, and value vectors, $d_m$ is the dimension of query vectors, $T$ denotes the matrix transpose operation. To enforce cross-modality interactions, we change the key and value vectors of the self-attention block to features in other modalities:
\begin{equation}
\small
    h_a = att(f_aW_Q, f_vW_K, f_vW_V),
\end{equation}
\begin{equation}
\small
    h_v = att(f_vW_Q, f_aW_K, f_aW_V),
\end{equation}
where $h_a$, $h_v$ are updated audio and visual features, $W_Q, W_K$, and $W_V$ are learnable parameters. We adopt the sharing parameter strategy for feature projection to reduce computation.

We adopt the MIL procedure under the weakly-supervised setting to obtain video-level scores. Unlike prior works, we process unimodal features individually to alleviate modality asynchrony. To be specific, fully-connected layers are used in each modality to generate unimodal logits. Then we take the summation of unimodal logits as the fused audio-visual logits while reserving the unimodal logits for the following contrastive learning. Finally, the top-\textit{K} audio-visual logits are average-pooled and put into a sigmoid activation to generate video-level scores for optimization. The entire procedure is formulated as:
\begin{equation}
\small
    l_a, l_v = W_af_a^{out}+b_a, W_vf_v^{out}+b_v
\end{equation}
\begin{equation}
\small
    p = \Theta(\Omega(\sigma(l_a \oplus l_v)))
\end{equation}
where $W_a, W_v, b_a, b_v$ are learnable parameters, $\Omega$ is the \textit{K}-max activation, $\sigma$ denotes the sigmoid function, $\oplus$ is the summation operation, $\Theta$ denotes average pooling, and $p$ is video-level prediction.

\subsection{MA-CIL}

To utilize more disengaged instances, we propose the MA-CIL module, which is shown on the right side of Figure~\ref{figure2}. Given the embeddings $h_a$, $h_v$, we perform unsupervised clustering to divide them into violent, normal, and background semi-bag representations based on the visual and audio logits. We argue that the discrepancy between semantic-irrelevant instances can be exploited to enrich model's capacity for discrimination.

To be specific, we first leverage the video-level probabilities $p$ to distinguish whether the given video contains violent events. In each mini-batch, for the video sequence $S_i$ that $p_i > 0.5$, top-\textit{K} instances with highest logits are clustered as the violence semi-bag $B_m^{vio}(i) = \{h_m(n)\}_{n=1}^{K_{vio}}, m\in\{a, v\}$. For the sequence $S_j$ that $p_j \leq 0.5$, top-\textit{K} instances are selected as the normal semi-bag $B_m^{nor}(j) = \{h_m(n)\}_{n=1}^{K_{nor}}, m\in\{a, v\}$. We hope adding contrast to the normal and violent events could help the model distinguish the violent extent of percepted signals.

Moreover, we argue that both normal and violent videos contain background snippets, and learning the difference between event-related segments and background noises could benefit the localization. Therefore, we select the bottom-\textit{K} instances of the whole mini-batch as the background semi-bag $B_m^{bgd} = \{h_m(n)\}_{n=1}^{K_{bgd}}, m\in\{a, v\}$. In each mini-batch, the model should contrast violent audio-visual instances against negative pairs constructed by violent instances and other instances (background and normal).

An intuitive way is to randomly pick intra- and inter-semi-bag instances in the opposite modality as positive and negative pairs. However, we argue that audio and visual violent instances with diverse positions could be semantically mismatched, such as expressing the beginning and ending of a violent event, respectively. Therefore, it is unnatural to assume that they share the same implication. In contrast, we conduct average pooling to embeddings of all violence instances in each bag and form a semi-bag-level representation $\mathcal{B}_m^{vio}, m\in\{a, v\}$. By doing so, the audio and visual representation both express event-level semantics, thereby alleviating the noise issue. To this end, we construct semi-bag-level positive pairs, which are assembled by audio and visual violent semi-bag representations $\mathcal{B}_a^{vio}, \mathcal{B}_v^{vio}$. We also construct semi-bag-to-instance negative pairs to maintain numerous contrastive samples, where violent semi-bag representations are combined with background and normal instance embeddings $h_m^{nor}, h_m^{bgd}, m\in \{a, v\}$ in the opposite modality as negative pairs.

We use the InfoNCE~\cite{van2018representation} as the training objective of this part, which closes the distance between positive pairs and enlarges the distance between negatives. The objective for audio violent semi-bag representation $B_a^{vio}(i)$ against visual normal instance embeddings $\{h_v^{nor}(n)\}_{n=1}^{K_{nor}}$ is formulated as:
\begin{equation}
\begin{split}
\small
    \mathcal{L}&_{ct}^{v2n}(B_a^{vio}(i)) = \\&-log \frac{e^{\phi(B_a^{vio}(i), B_v^{vio}(i))/\tau)}}{e^{\phi(B_a^{vio}(i), B_v^{vio}(i))/\tau)}+\sum_{n=1}^{K_{nor}}e^{\phi(B_a^{vio}(i), h_v^{nor}(n)/\tau)}},
\end{split}
\end{equation}
where $\phi$ denotes cosine similarity function, $\tau$ is the temperature hyperparameter, $K_{nor}$ denotes the normal instances number in the whole mini-batch. Similarly, the objective for audio violent semi-bag representation $B_a^{vio}(i)$ against visual background instances embeddings $\{h_v^{bgd}(n)\}_{n=1}^{K_{bgd}}$ is formulated as:
\begin{equation}
\begin{split}
\small
    \mathcal{L}&_{ct}^{v2b}(B_a^{vio}(i)) = \\&-log \frac{e^{\phi(B_a^{vio}(i), B_v^{vio}(i))/\tau)}}{e^{\phi(B_a^{vio}(i), B_v^{vio}(i))/\tau)}+\sum_{n=1}^{K_{bgd}}e^{\phi(B_a^{vio}(i), h_v^{bgd}(n)/\tau)}},
\end{split}
\end{equation}
where $K_{bgd}$ denotes the background instance number in the whole mini-batch. The visual-against-audio counterparts are highly similar, thus we omit these for concise writing.

\subsection{Self-Distillation}
The audio-visual interactions provided by the former parts could introduce abundant modality noises, and modality asynchrony also results in the semantic mismatch of multimodal and unimodal features in the same temporal position. To address these issues, we argue that training a similar visual network simultaneously enables the model to ensemble unimodal and multimodal knowledge. With a controllable co-distillation strategy, our proposed module warrants modality noise reduction and robust modality-agnostic knowledge.

Specifically, we propose an analogous unimodal network that contains comparable architecture with our two-stream network. The cross-modality attention block is substituted by the standard transformer encoder block including self-attention. During training, the unimodal network is trained with a relatively small learning rate, and parameters of the same layers are infused into the audio-visual network with an exponential moving average strategy:
\begin{equation}
\small
    \theta_{av} \gets m\theta_{av} + (1-m)\theta_v 
\end{equation}
where $\theta_{av}$ and $\theta_{v}$ denotes parameters of the audio-visual model and visual model, respectively, $m$ denotes the control hyperparameter following a cosine scheduler that increases from the original value $\hat{m}$ to 1 during training.

\subsection{Learning Objective}

The entire framework is optimized in a joint-training manner. For the video-level prediction $p$, we leverage binary cross-entropy $\mathcal{L}_\mathcal{B}$ as the training objective and use a linearly growing strategy to control the weight of contrastive loss. The total objective is:
\begin{equation}
\begin{split}
\small
    \mathcal{L}_{av} = &\frac{\lambda_{v2n}(t)}{K_{vio}}\sum_{i}
    (\mathcal{L}_{ct}^{v2n}(B_a^{vio}(i)) + \mathcal{L}_{ct}^{v2n}(B_v^{vio}(i))) + \\&    \frac{\lambda_{v2b}(t)}{K_{vio}}\sum_{i}
    (\mathcal{L}_{ct}^{v2b}(B_a^{vio}(i)) + (\mathcal{L}_{ct}^{v2b}(B_v^{vio}(i))) + \mathcal{L}_{B}
\end{split}
\end{equation}
\begin{equation}
\small
    \lambda(t) = min(r*t, \Lambda)
\end{equation}
where $K_{vio}$ denotes the number of violence semi-bags in the whole mini-batch, $\lambda(t)$ is a controller to increase weight within a few epochs linearly, $r$ denotes the growing ratio, $t$ is the current epoch, and $\Lambda$ denotes the maximum weight. 

The visual network is optimized via the BCE loss with video-level labels to distill unimodal knowledge. The two objectives are optimized simultaneously during training while in the inference phase, only the audio-visual network is used for prediction.

\section{Experiment}
We design experiments to verify our model from two perspectives, the end-to-end framework compared with state-of-the-art methods and assembling with other networks as plug-in modules. Experimental details and analyses are introduced as follows.

\subsection{Dataset and Evaluation Metric}

\textbf{XD-Violence}~\cite{wu2020not} dataset is by far the only available large-scale audio-visual dataset for violence detection, which is also the largest dataset compared with other unimodal datasets. XD-Violence consists of 4,757 untrimmed videos (217 hours) and six types of violent events, which are curated from real-life movies and in-the-wild scenes on YouTube. Although previous methods adopt some popular datasets~\cite{sultani2018real, liu2018future} as benchmarks, we argue that these datasets only contain unimodal visual contents, which cannot perform cross-modality interactions and further verify our proposed multimodal framework. Hence, following~\cite{wu2020not, pang2021violence}, we select the large-scale audio-visual dataset XD-Violence as benchmark. During inference, we utilize the frame-level average precision (AP) as evaluation metrics following previous works~\cite{wu2020not, pang2021violence, tian2021weakly}.

\subsection{Implementation Details}

To make a fair comparison, we adopt the same feature extracting procedure as prior methods~\cite{wu2020not, pang2021violence, wu2021learning, tian2021weakly}. Concretely, we use the I3D~\cite{carreira2017quo} network pretrained on the Kinetics-400 dataset to extract visual features. Audio features are extracted via the VGGish~\cite{gemmeke2017audio, hershey2017cnn} network pretrained on a large YouTube dataset. The visual sample rate is set to be 24 fps, and visual features are extracted by a sliding window with a size of 16 frames. For the auditory data, we first divide each audio into 960-ms overlapped segments and compute the log-mel spectrogram with 96 $\times$ 64 bins. 

The entire network is trained on an NVIDIA Tesla V100 GPU for 50 epochs. We set the batch size as 128 and the initial learning rate as 4e-4, which is dynamically adjusted by a cosine annealing scheduler. For the visual distillation network, the learning rate is set as 8e-5. We use Adam~\cite{kingma2014adam} as the optimizer without weight decay. During optimization, the weighted hyperparameter $r, \Lambda_{v2b}, \Lambda_{v2n}$ are 0.1, 1.5, and 1.5, respectively. The initial distillation weight $\hat{m}$ is set to 0.91. The temperature $\tau$ of InfoNCE~\cite{van2018representation} is set to be 0.1. The hidden dimension of our two-stream network is 128, and the dropout rate is 0.1. For the MIL, we set the value $K$ of \textit{K}-max activation as $\left \lfloor \frac{T}{16} + 1 \right \rfloor$, where $T$ denotes the length of input feature.

\subsection{Comparisons with State-of-the-Arts}

\begin{table}[tb]
\caption{Comparison of the frame-level AP performance with unsupervised and weakly-supervised baselines. $\dagger$ denotes results re-implemented by integrating logits of two identical networks with audio and visual inputs, and * indicates re-implemented by fusing audio and visual features as inputs.}
\begin{tabular}{@{}clccc@{}}
\toprule
 Manner                            & Method                                            & Modality       & AP (\%)        & Param.       \\ \midrule
\multirow{3}{*}{Unsup.}            & SVM baseline                                      & V              & 50.78          &   /           \\
                                   & OCSVM~\cite{scholkopf1999support}                 & V              & 27.25          &   /           \\
                                   & Hasan et al.~\cite{hasan2016learning}             & V              & 30.77          &   /           \\ \midrule
\multirow{11}{*}{W. Sup.}          & Sultani et al.~\cite{sultani2018real}             & V              & 73.20          &   /          \\
                                   & Wu et al.~\cite{wu2021learning}                   & V              & 75.90          &   /           \\
                                   & RTFM~\cite{tian2021weakly}                        & V              & 77.81          &  12.067M            \\
                                   & RTFM*~\cite{tian2021weakly}                       & A+V            & 78.10          &  13.510M            \\
                                   & RTFM$\dagger$~\cite{tian2021weakly}               & A+V            & 78.54          &  13.190M           \\ 
                                   & Li et al.~\cite{li2022self}                       & V              & 78.28          &  /            \\
                                   & Wu et al.~\cite{wu2020not}                        & A+V            & 78.64          &  0.843M      \\
                                   & Wu et al.$\dagger$~\cite{wu2020not}               & A+V            & 78.66          &  1.539M      \\
                                   & Pang et al.~\cite{pang2021violence}               & A+V            & 81.69          &  1.876M      \\ \cmidrule(l){2-5} 
                                   & Ours (light)                                      & A+V            & \textbf{82.17} &  \textbf{0.347M}      \\ 
                                   & Ours (full)                                       & A+V            & \textbf{83.40} &  \textbf{0.678M}      \\ \bottomrule
\end{tabular}
\label{exp:table1}
\end{table}

We compare our proposed approach with state-of-the-art models, including (1) unsupervised methods: SVM baseline, OCSVM~\cite{scholkopf1999support}, and Hasan et al.~\cite{hasan2016learning}; (2) unimodal weakly-supervised methods: Sultani et al.~\cite{sultani2018real}, RTFM~\cite{tian2021weakly}, Li et al.~\cite{li2022self}, and Wu et al.~\cite{wu2021learning}; (3) audio-visual weakly-supervised methods: Wu et al.~\cite{wu2020not} and Pang et al.~\cite{pang2021violence}. We report the AP results on XD-Violence dataset in Table~\ref{exp:table1}. 

With video-level supervisory signals, our method outperforms all previous unsupervised approaches by a large margin. Moreover, compared with previous unimodal weakly-supervised methods, our model surpasses prior results with a minimum of 5.12\%, showing the necessity of utilizing multimodal cues for violent detection. 

To further demonstrate the efficacy of our modality-aware contrastive instance learning and cross-modality distillation, we select state-of-the-art methods~\cite{wu2020not, pang2021violence} as audio-visual baselines and re-implement SOTA unimodal MIL method~\cite{tian2021weakly} with two modality-expansion strategies. First, following~\cite{wu2020not}, we fuse the audio and visual features in an early way as model inputs. This approach forbids the intermediate modality interaction in the network, aiming to show the performance of simply integrating multimodal data. Considering some networks may be unsuitable for multimodal inputs, we put forward another strategy to train two unimodal networks simultaneously and generate audio and visual logits, respectively. The audio-visual predictions are generated by fusing unimodal logits. Results show that our framework achieves 1.71\% higher performance against state-of-the-art method Pang et al.~\cite{pang2021violence}, which verifies that our MA-CIL and SD modules are practical for violence detection. Our method outperforms RTFM* and Wu et al. by 5.30\% and 4.76\% for multimodal variants using audio-visual inputs. For variants using two-stream architecture, we observe that our model surpasses RTFM$\dagger$ and Wu et al.$\dagger$ by 4.86\% and 4.74\%, respectively, which suggests that modality-aware interactions are indispensable for multimodal scenarios. To conclude, using the same input features, our method achieves superior performance compared with all audio-visual methods, showing the effectiveness of our entire proposed audio-visual framework.

\subsection{Plug-in Module}

\begin{table}[tb]
\caption{Results on proposed MA-CIL and SD modules as plug-in modules. * indicates results re-implemented by fusing audio and visual features as inputs. $\dagger$ denotes re-implemented by integrating logits of two identical networks with audio and visual inputs, respectively. $\ddagger$ is the ablated model that removes the fusion module and mutual loss.}
\begin{tabular}{lcccc}
\hline
Method                                          & MA-CIL                & SD                & AP (\%)                                   & Param.   \\ \hline
Wu et al.~\cite{wu2020not}                      & \XSolidBrush          & \XSolidBrush      & 78.64                                     & 0.843M   \\
Wu et al.$\dagger$~\cite{wu2020not}             & \XSolidBrush          & \XSolidBrush      & 78.66                                     & 1.539M   \\
Wu et al.~\cite{wu2020not}                      & \XSolidBrush          & \Checkmark        & 80.07 \color{teal}{(1.43$\uparrow$)}      & 1.612M   \\
Wu et al.$\dagger$~\cite{wu2020not}             & \Checkmark            & \XSolidBrush      & 79.98 \color{teal}{(1.32$\uparrow$)}      & 1.539M   \\ \hline
RTFM~\cite{tian2021weakly}                      & \XSolidBrush          & \XSolidBrush      & 77.81                                     & 12.067M  \\
RTFM*~\cite{tian2021weakly}                     & \XSolidBrush          & \XSolidBrush      & 78.10                                     & 13.510M  \\
RTFM$\dagger$~\cite{tian2021weakly}             & \XSolidBrush          & \XSolidBrush      & 78.54                                     & 13.190M  \\
RTFM*~\cite{tian2021weakly}                     & \XSolidBrush          & \Checkmark        & 80.40 \color{teal}{(2.30$\uparrow$)}      & 25.577M  \\
RTFM$\dagger$~\cite{tian2021weakly}             & \Checkmark            & \XSolidBrush      & 80.00 \color{teal}{(1.46$\uparrow$)}      & 13.190M  \\ \hline
Pang et al.~\cite{peng2020learning}             & \XSolidBrush          & \XSolidBrush      & 81.69                                     & 1.876M   \\
Pang et al.$\ddagger$~\cite{peng2020learning}   & \XSolidBrush          & \XSolidBrush      & 80.03                                     & 1.086M   \\
Pang et al.~\cite{peng2020learning}             & \XSolidBrush          & \Checkmark        & 81.21 \color{teal}{(1.18$\uparrow$)}      & 2.138M   \\
Pang et al.~\cite{peng2020learning}             & \Checkmark            & \XSolidBrush      & 80.90 \color{teal}{(0.87$\uparrow$)}      & 1.086M   \\
Pang et al.~\cite{peng2020learning}             & \Checkmark            & \Checkmark        & 82.21 \color{teal}{(2.18$\uparrow$)}      & 1.613M   \\ \hline
\end{tabular}
\label{exp:table2}
\end{table}

We also argue that our proposed modules have satisfying generalizability and are capable of enhancing other networks. To this end, we combine our framework with state-of-the-art methods and evaluate the performance. First, we re-implement the state-of-the-art audio-visual method~\cite{pang2021violence} using the official implementations provided by the original paper. Then we select the unimodal method with publicly available codes RTFM~\cite{tian2021weakly} as the unimodal baseline, which is ameliorated to multimodal networks by two means we mentioned above (* and $\dagger$). For the multimodal method Wu et al.~\cite{wu2020not}, we use the two-stream variant to examine the performance of our MA-CIL module and use the native version for combining with SD. Since the unimodal network RTFM~\cite{tian2021weakly} and the audio-visual method Wu et al.~\cite{wu2020not} can only be amalgamated with MA-CIL in the two-stream network manner ($\dagger$), while SD should be assembled in an early modality fusion way (*), we can only combine these frameworks with our modules separately. For the multimodal approach~\cite{pang2021violence}, we both testify the joint and independent enhancement performances of our MA-CIL and SD modules.

\begin{table}[t]
\caption{Ablation studies on different components of our proposed framework.}
\setlength{\tabcolsep}{3.5mm}{
\begin{tabular}{@{}ccccc@{}}
\toprule
 Index      & Two-Stream            & MA-CIL                 & SD                & AP (\%)               \\ \midrule
 1          & \Checkmark            & \XSolidBrush           & \XSolidBrush      & 71.37                 \\
 2          & \Checkmark            & \XSolidBrush           & \Checkmark        & 74.01                 \\
 3          & \Checkmark            & \Checkmark             & \XSolidBrush      & 82.17                 \\ \midrule
 4          & \Checkmark            & \Checkmark             & \Checkmark        & \textbf{83.40}        \\ \bottomrule
\end{tabular}
\label{exp:table3}}
\end{table}

We report the results on the XD-Violence dataset in Table~\ref{exp:table2}. First, we observe that MA-CIL boosts the unimodal baselines Wu et al.~\cite{wu2020not} and RTFM~\cite{tian2021weakly} for 1.32\% and 1.46\%, respectively, showing that our contrastive learning method improves the discrimination of models. We also note that equipped with the SD module, the performances of~\cite{wu2020not, tian2021weakly} also gain an increase of 1.43\% and 2.30\%. For the multimodal baseline~\cite{pang2021violence}, we remove the mutual loss and multimodal fusion modules and leverage the vanilla attention-based variant ($\ddagger$) for comparison. Results show that enhanced with MA-CIL and SD separately or jointly both achieve accuracy boosts. In summary, we conclude that integrating our MA-CIL and SD modules is beneficial to numerous networks and our modules can be utilized flexibly depending on specific usages. 

\subsection{Complexity Analysis}

As we mentioned before, we propose a computation-friendly framework that does not introduce too many parameters. To support our claims, we compare parameter amounts with previous methods, which are shown in the Param. column of Table~\ref{exp:table1},~\ref{exp:table2}. In Table~\ref{exp:table1}, we report the parameter amounts of previous works we re-implement and our proposed framework, where Ours (light) denotes the ablated model without self-distillation, and Ours (full) indicates the full model with MA-CIL and SD. In Table~\ref{exp:table2}, we provide parameter amounts of the raw methods and our enhancement variants.

From the comparison with other methods, we observe that Ours (light) holds the smallest model size (0.347M) while outperforming all previous methods. Combined with the SD module, our full model still has fewer parameter amounts and achieves the best performance. This result demonstrates the efficiency of our framework, which leverages a much simpler network yet gains better performance. As shown in Table~\ref{exp:table2}, we note that the MA-CIL method does not include any parameters, which exploits the intrinsic prior of multimodal instances and obtains model discrimination with no computation cost. When boosting the multimodal model~\cite{pang2021violence}, the enhanced model has comparable size to the raw model due to the analogous model structure. This suggests that our proposed modules are flexible to be adapted to multimodal networks.

\begin{table}[htbp]
\caption{Ablation study for the hyperparameters in the proposed modality-aware contrastive instance learning.}
\setlength{\tabcolsep}{4mm}{
\begin{tabular}{@{}ccccc@{}}
\toprule
    Index &     $\Lambda_{v2a}$      & $\Lambda_{v2b}$      & ratio ($r$)   & AP (\%)              \\ \midrule
    1  &       \multirow{3}{*}{1.0}  & \multirow{3}{*}{1.0} & 0.1           & 82.62                \\
    2  &                             &                      & 0.3           & 82.67                \\ 
    3  &                             &                      & 3.0           & 82.09                \\ \midrule
    4  &        \multirow{3}{*}{1.5} & \multirow{3}{*}{1.0} & 0.1           & 82.95                \\
    5  &                             &                      & 0.3           & 81.37                \\ 
    6  &                             &                      & 3.0           & 82.15                \\ \midrule
    7  &        \multirow{3}{*}{1.0} & \multirow{3}{*}{1.5} & 0.1           & 83.21                \\
    8  &                             &                      & 0.3           & 82.62                \\
    9  &                             &                      & 3.0           & 81.68                \\ \midrule
    10 &        \multirow{3}{*}{1.5} & \multirow{3}{*}{1.5} & 0.1           & \textbf{83.40}       \\
    11 &                             &                      & 0.3           & 81.61                \\
    12 &                             &                      & 3.0           & 82.14                \\ 
    \bottomrule
\end{tabular}
\label{exp:table4}}
\end{table}

\subsection{Ablation Studies}

To further investigate the contribution of our proposed modules, we conduct ablation experiments to demonstrate how each aspect of our framework affects the overall performance.

We first conduct experiments on the effectiveness of each component, and the results are shown in Table~\ref{exp:table3}. The vanilla two-stream network without MA-CIL and SD achieves a performance of 71.37\%. We argue that the limited performance is driven by the small-scale model architecture. Equipped with MA-CIL, we observe a remarkable performance boost from 71.37\% to 82.17\%, proving that our proposed contrastive method benefits model discrimination and further improves the detection performance. We then investigate the role of our SD module. Combining our SD module to the raw two-stream network and network with MA-CIL, the ablated models achieve the AP increase of 2.64\% and 1.23\%, respectively. This indicates that the SD module is effective both with and without contrastive learning, and the two modules complement each other for a better violence detection performance.

\begin{figure}[tb]
\centering
\includegraphics[width=0.45\textwidth]{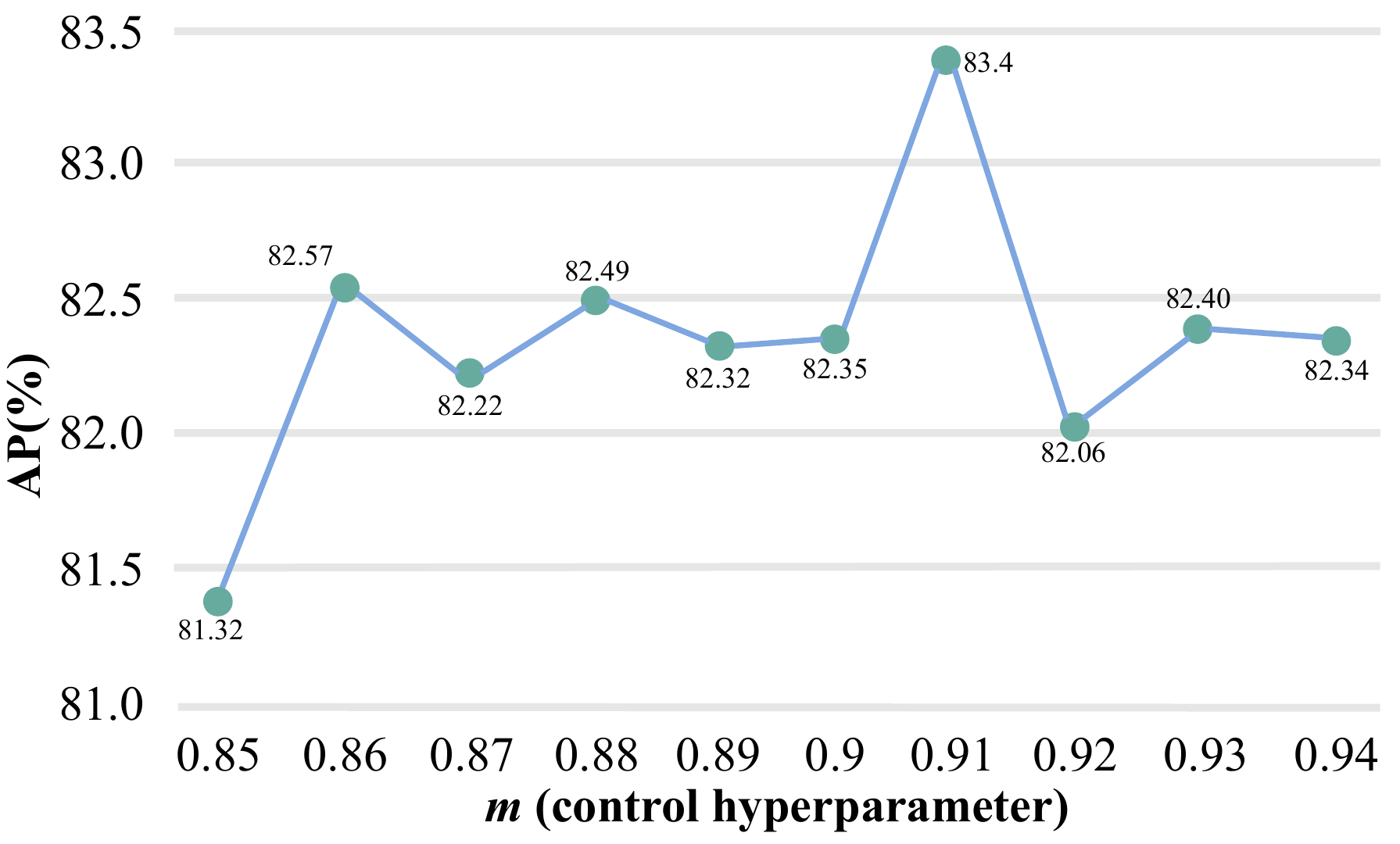}
\vspace{-2mm}
\caption{Ablation studies of different settings for control hyperparameter $m$ in our self-distillation module.}
\vspace{-2mm}
\label{figure3}
\end{figure}

Then we perform ablation studies on the loss control strategy of our modality-aware contrastive instance learning. As shown in Table~\ref{exp:table4}, $\Lambda_{v2b}$, $\Lambda_{v2n}$ denote the maximum weights of $\mathcal{L}_{ct}^{v2b}$, $\mathcal{L}_{ct}^{v2n}$, respectively. $r$ is the linearly increasing ratio. Table~\ref{exp:table4} shows the results of different settings about $\Lambda_{v2b}$, $\Lambda_{v2n}$, and $\tau$. We observe that the optimal setting is $\Lambda_{v2b}$ = 1.5, $\Lambda_{v2n}$ = 1.5, $\tau$ = 0.1, while training with the full weights from the very beginning (r=3.0) brings worse performance. This suggests that gently raising the proportion of contrastive loss is a plausible training strategy, where the model focuses more on the quality of audio and visual embeddings in the early stage and learning feature discrimination afterwards.

Finally, we investigate the control hyperparameter $m$ of the self-distillation block in our proposed method as shown in Figure~\ref{figure3}. Results show that the best performance achieves at $m$ = 0.91.

\subsection{Qualitative Analysis}

\begin{figure}[tb]
\centering
\includegraphics[width=0.45\textwidth]{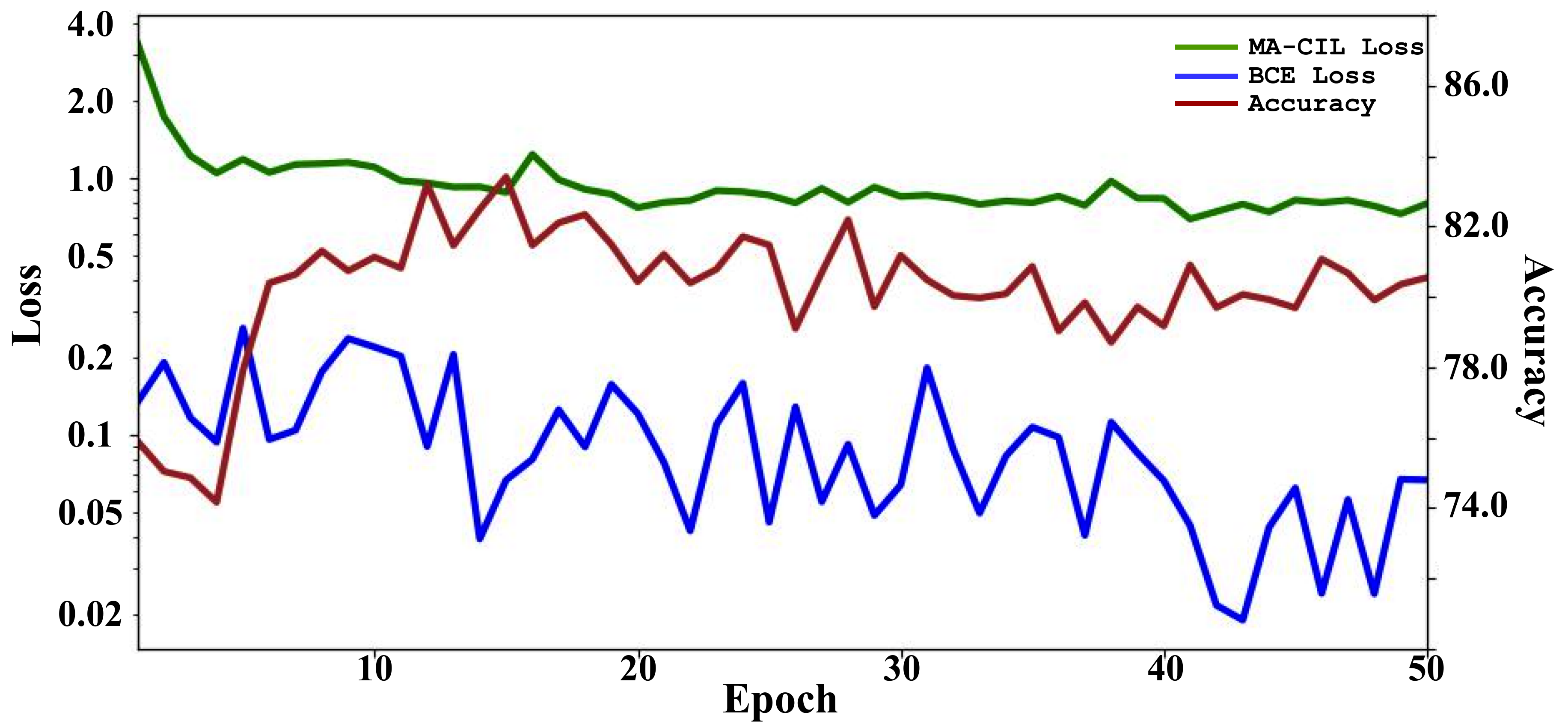}
\vspace{-2mm}
\caption{Illustration of the accuracy and loss curves in 50 epochs during training. The red curve denotes the video-level prediction accuracy. The ranges of BCE loss and contrastive loss are shown in blue and green curves, respectively.}
\vspace{-2mm}
\label{figure4}
\end{figure}

\begin{figure}[tb]
\centering
\includegraphics[width=0.45\textwidth]{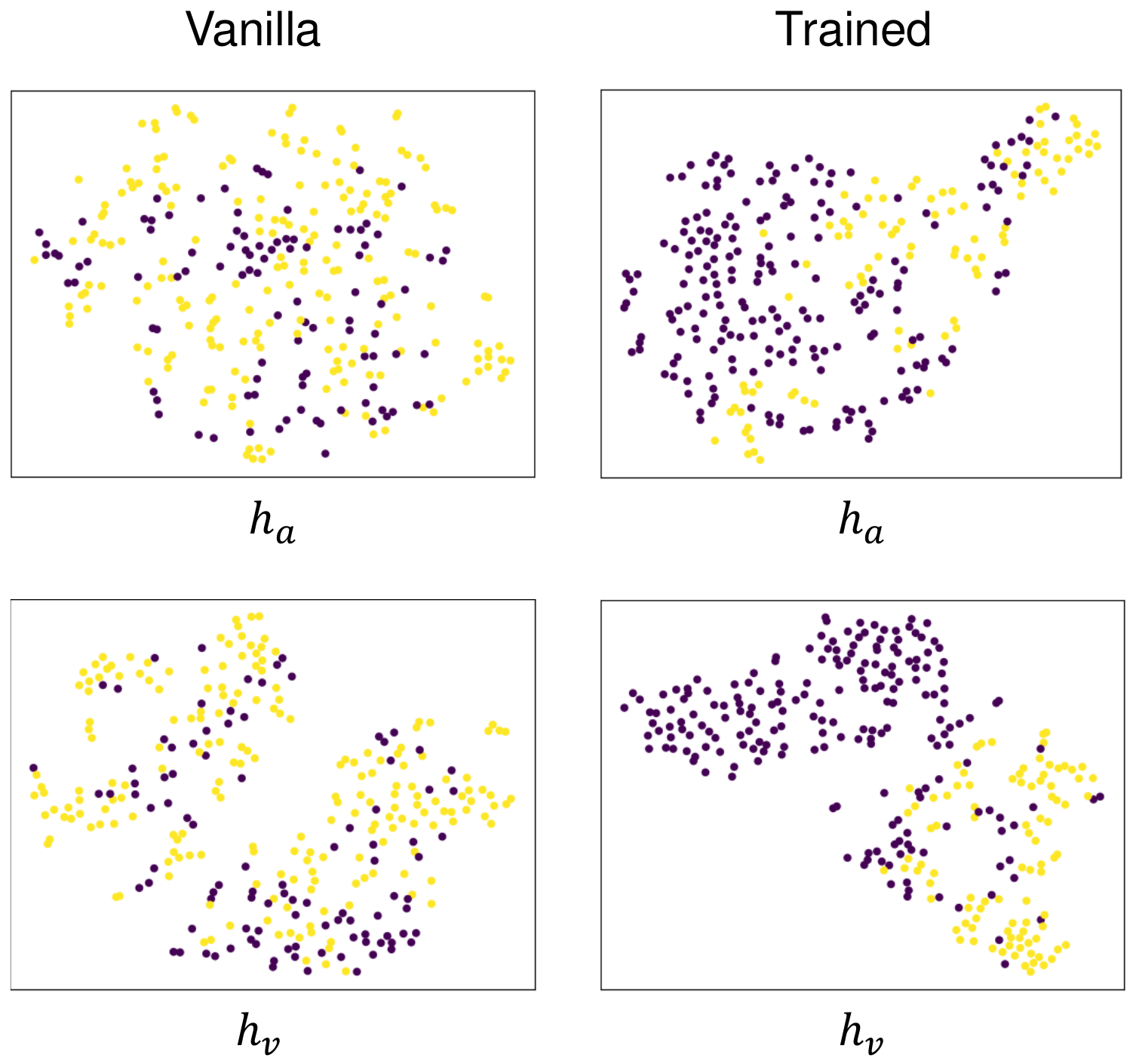}
\vspace{-2mm}
\caption{Feature space visualizations of the vanilla features and the output of our model on XD-Violence testing videos.}
\vspace{-2mm}
\label{figure5}
\end{figure}

\begin{figure*}[htbp]
\centering
\includegraphics[width=0.94\textwidth]{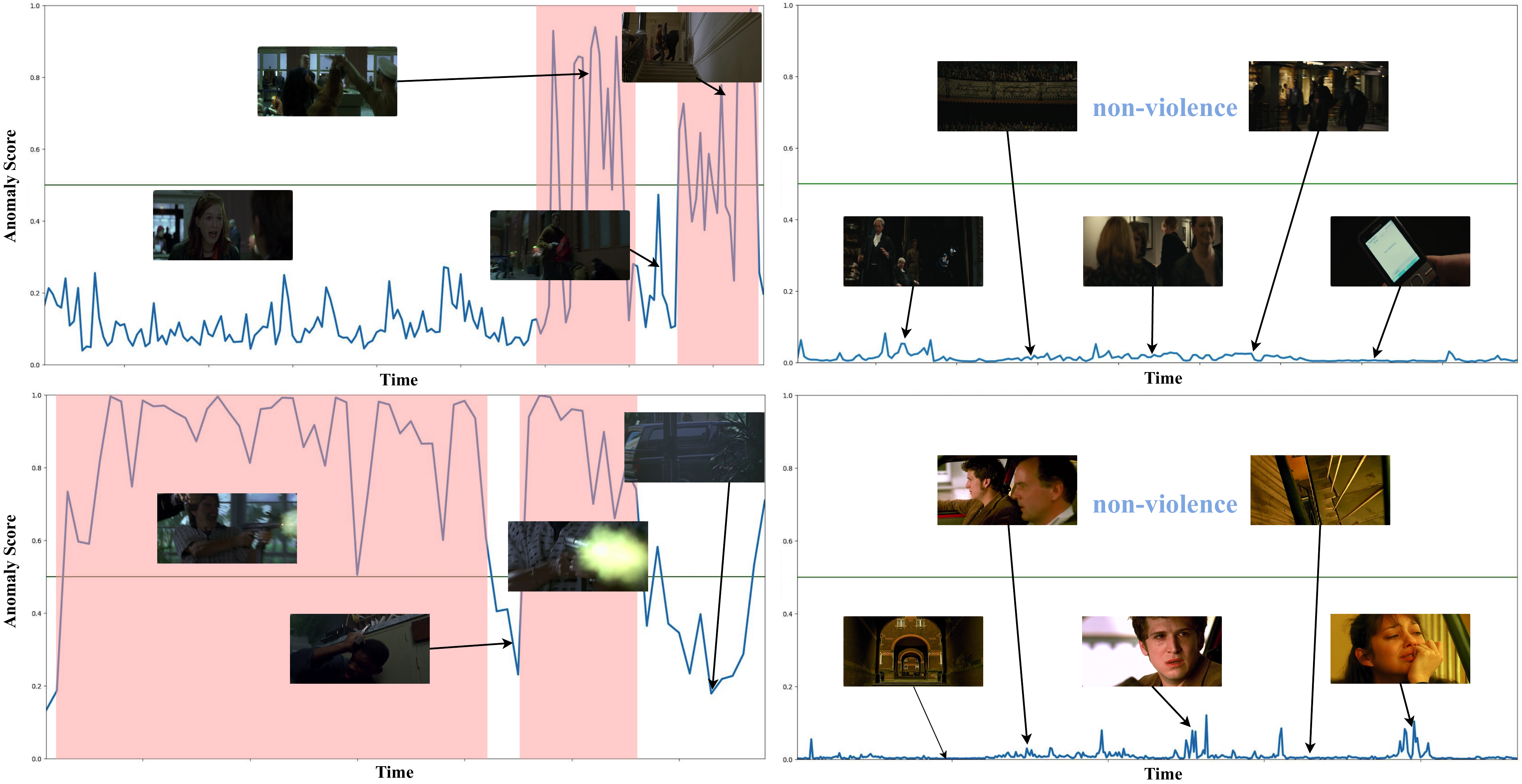}
\vspace{-2mm}
\caption{Visualization of results on the XD-Violence test set. Red regions are the temporal ground-truths of violent events.}
\vspace{-3mm}
\label{figure6}
\end{figure*}

We first visualize the variation of the training loss and video-level accuracy on the XD-Violence dataset. Results are shown in Figure~\ref{figure4}, where the red curve denotes the video-level accuracy, and the blue and green curves denote BCE loss and contrastive loss, respectively. For the prediction accuracy, we observe a sudden decrease in the first 10 training epochs, where the contrastive learning constraints are gradually applied with the increasing weights. After learning the discrimination for a few epochs, the training accuracy begins to increase and finally outperforms previous results. A similar conclusion also appears in the loss curves. The reduction of the BCE loss comes from the first few epochs, where the model is required to generate high-quality embeddings. The contrastive loss has a lasting decline in dozens of epochs, which means the constraints enforce the model to differentiate instances for a long training period. These curves also denote that the two objectives are co-optimized without interfering with each other. We argue that contrastive learning plays a complementary role to traditional MIL learning, and this insight further demonstrates the generalizability of our methods.

We also provide t-SNE~\cite{van2008visualizing} visualizations about the distributions of audio and visual features on the XD-Violence test set. Results are shown in Figure~\ref{figure5}, where yellow dots denote background segments and purple dots are violent features. We can find that the violent and non-violent features are clearly clustered, and the distance between uncorrelated features is enlarged after the training procedure. This reveals that aided by our proposed network, instances are successfully differentiated in both audio and visual modalities, further indicating the effectiveness of our proposed framework.

Finally, we provide visualizations of prediction results presented in Figure~\ref{figure6}. Our model accurately localizes the anomalous events and even identifies normal events of a very short duration between two violent events. In non-violent videos, the magnitudes between normal and background segments are also evident. Scores of normal events will be a little higher than the background segments yet far less than the violent segments, and our method generates nearly zero predictions for the background snippets. These results show that our proposed approach enables the model to perceive the discrepancy between segments in different types (violent, normal, and background), and further contribute to the violent detection.

\section{Conclusion}
In this paper, we investigate the model asynchrony and undifferentiated instances phenomena of MIL under audio-visual scenarios, and further show the impact on weakly-supervised audio-visual learning. Then a modality-aware contrastive instance learning with a self-distillation framework is proposed to address these issues. To be specific, we design a lightweight two-stream network to generate audio and visual embedding and logits. Furthermore, a cross-modality contrast is applied to audio and visual instances of different semantics, which involves more unused instances for better discrimination and alleviates the modality inconsistency. To diminish training noises, a self-distillation module is leveraged to transfer visual knowledge to the audio-visual network, by which the semantic gaps between unimodal and multimodal features are narrowed. Our framework outperforms previous methods on the XD-Violence dataset with minor expenses. Besides, assembled with our contrastive learning and self-distillation modules, several prior methods achieve higher detection accuracy, showing the capability as plug-in modules to ameliorate other networks.

\begin{acks}
This work was supported by National Natural Science Foundation of China (No. 62172101, No. 61976057). This work was supported (in part) by the Science and Technology Commission of Shanghai Municipality (No. 21511101000, No. 21511100602), and the SPMI Innovation and Technology Fund Projects (SAST2020-110).

\end{acks}

\clearpage
\bibliographystyle{ACM-Reference-Format}
\bibliography{reference}

\end{document}